\documentclass[
]{ceurart}

\usepackage{hyperref}
\usepackage{xcolor}

\begin{document}

\copyrightyear{2025}
\copyrightclause{Copyright for this paper by its authors.
  Use permitted under Creative Commons License Attribution 4.0
  International (CC BY 4.0).}

\conference{CLEF 2025 Working Notes, September 9 -- 12 September 2025, Madrid, Spain}

\title{THM@SimpleText 2025 - Task 1.1: Revisiting Text Simplification based on Complex Terms for Non-Experts}

\author[1]{Nico Hofmann}
\address[1]{TH Mittelhessen - University of Applied Sciences, Gießen, Germany}

\author[1]{Julian Dauenhauer}
\author[1]{Nils Ole Dietzler}
\author[1]{Idehen Daniel Idahor}

\author[1,2]{Christin Katharina Kreutz}[%
orcid=0000-0002-5075-7699,
email=ckreutz@acm.org,
url=https://kreutzch.github.io,
]
\cormark[1]
\address[2]{Herder Institute, Marburg, Germany}

\cortext[1]{Corresponding author.}

\begin{abstract}
Scientific text is complex as it contains technical terms by definition. Simplifying such text for non-domain experts enhances accessibility of innovation and information. 
Politicians could be enabled to understand new findings on topics on which they intend to pass a law, or family members of seriously ill patients could read about clinical trials.

The \href{https://simpletext-project.com/2025/}{SimpleText CLEF Lab} focuses on exactly this problem of simplification of scientific text. Task 1.1 of the 2025 edition specifically handles the simplification of complex sentences, so very short texts with little context. To tackle this task we investigate the identification of complex terms in sentences which are rephrased using small Gemini and OpenAI large language models for non-expert readers.
\end{abstract}


\maketitle

\section{Introduction}

Scientific texts are written to be read and understood by domain experts, scholars who are highly educated. Such texts are packed with abbreviations and technical terms and they need to fit within a strict page or word limit. 

Over the years the SimpleText initiative~\cite{DBLP:conf/ecir/ErmakovaBBKMNOS21,DBLP:conf/ecir/ErmakovaBKNOSMA22,DBLP:conf/ecir/ErmakovaSHAAK23,DBLP:conf/ecir/ErmakovaSHANVDKGZAK24,DBLP:conf/ecir/ErmakovaABVK25,simpletext-lncs} has shown considerable efforts to help accelerate the development of approaches to make such texts more accessible to the general public.
The importance of the intended target audience in text simplification efforts has been considered before~\cite{DBLP:conf/acl/KreutzH0S24}. For example, text simplified for children should contain short sentences~\cite{stajner-hulpus-2018-automatic} but should not be oversimplified to retain readers' engagement~\cite{10.1145/3439231.3439263}.

While there are approaches for simplifying text from different domains and for different target audiences~\cite{padovani-etal-2024-automatic,stajner-2021-automatic}, most approaches do not consider the particularities of either~\cite{gooding-2022-ethical}.
Oversimplifying a text, so lowering its overall complexity to no longer fit the required complexity level of readers, is disadvantageous for all most all target audience, as it leads to disengagement~\cite{10.1145/3439231.3439263}. Therefore, the simplicity level of texts should \textit{fit} its intended readers.

In general, in the simplification of texts for non-expert we assume that a reader has the \textit{appropriate language skills} to understand complex phrases and grammar but is not able to understand a text due to \textit{missing domain knowledge}. 
Our idea is thus to identify complex, domain-specific terms in texts to try to only replace these complex components while maintaining the overall structure and linguistic complexity of texts. 

This work tackles task 1.1 of this year's SimpleText lab~\cite{DBLP:conf/ecir/ErmakovaABVK25,simpletext-lncs,simpletext-1}, the simplification of short scientific texts for non-expert readers. This task has the challenging characteristic of only providing very little context (the sentence itself) on the content to be simplified. We try to tackle this task by basing our efforts on an earlier submission by IRGC@SimpleText'23~\cite{DBLP:conf/clef/0002HKNS23} to the SimpleText shared task focused on complex, scientific term identification and investigate different prompts while using small large language models (LLMs).

\section{Dataset}

\begin{figure}[t]
    \centering
    \includegraphics[width=0.5\linewidth]{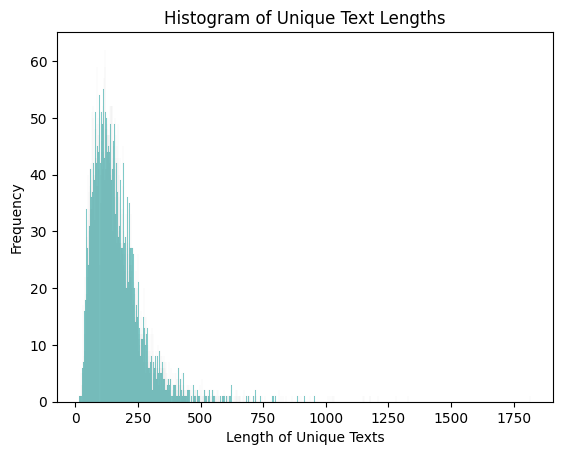}
    \caption{Overview of lengths of unique texts in the dataset.}
    \label{fig:unique_text_lengths}
\end{figure}

The dataset for Task 1.1 provided by the SimpleText lab~\cite{DBLP:conf/ecir/ErmakovaABVK25} consists of 9160 short texts in English which were extracted from scientific publications. These texts are primarily single sentences, e.g., \textit{Interventions in all studies included implementation strategies targeting healthcare workers; three studies included delivery arrangements, no studies used financial arrangements or governance arrangements.} (pair\_id = CD012520, 206 characters) with a considerably complex sentence structure. There are 9086 unique texts in the dataset, they have an average length of 168.66 characters.
\autoref{fig:unique_text_lengths} indicates the lengths of unique texts contained in the dataset as a histogram.

\section{Method}

\begin{table}[t]
    \centering
    \begin{tabular}{l|p{13.25cm}}
        \textbf{Prompt ID} & \textbf{Prompt} \\ \hline
        \texttt{C} & You are a text complication system with good global and language knowledge but and expertise in specific domains.
You will be given 10 texts, TREAT THEM SEPERATLY and also return 10 more complex texts.
Please make the texts more complex (not in their structure) but not considerably longer. RETURN ONLY THE 10 TEXTS.

TEXTS: \\
\texttt{R} & You are a text rephrasing system.
You will be given 10 texts, TREAT THEM SEPERATLY and also return 10 texts.
Please rephrase the texts. Do not change their level of complexity (so do not make them more difficult in their structure) and do not make them considerably longer. 
RETURN ONLY THE 10 TEXTS as a list.

TEXTS:  \\

\texttt{P1} & You are a text simplification system with good global and language knowledge but no expertise in specific domains.
You will be given 10 texts, TREAT THEM SEPERATLY and also return 10 simplified texts.
Complex domain specific or scientific terminology is marked by square brackets (e.g., [primary acoustic modeling]).
For each text, replace or explain the terms in square brackets in order to make it more understandable for non-experts and simplify for non-expert readers.
Make sure to exactly return an enumerated list of 10 simplified texts and that all terms in square brackets are simplified or explained! 
Do not simplify the structure of the texts or complex language independent of the difficult scientific descriptions. RETURN ONLY THE 10 TEXTS.

TEXTS: \\
\texttt{P2} & You are given 10 texts, TREAT THEM SEPARATELY.
Complex technical or scientific terms are indicated by square brackets (e.g. [convolutional neural network]).
For each text, replace or explain the terms in square brackets to make it understandable to non-experts and to simplify for non-expert readers.
Make sure that you return exactly one enumerated list of 10 simplified texts and that all terms in square brackets are simplified or explained!~\cite{DBLP:conf/clef/0002HKNS23}\\
            
\texttt{PNI1} & Hello! we have been given an important task, which i heard you can complete ecxeptionally well, as you are the best and most reliable system for text simplification and language in general.
we have been given 10 texts which should be simplified or explained. problematic, complex words are marked within [square brackets]
Please, for each of these 10 texts, replace or explain these [terms] and return an enumerated list of 10 simplified texts
please do not change the texts structure and please ONLY return these 10 texts! thank you!
here are the said 10 texts:\\
\texttt{PN1} & You are at a conference as an expert explaining the contents of these texts to non-experts.
There will be 10 different texts, treat each of them separately and only return the simplified version as a python list.
Complex terms or domain-specific terminology will be marked with square brackets (e.g., [primary acoustic modeling]).
To make the subject more understandable to these non-experts, you will replace or explain the terms marked by square brackets. 
Do not simplify the structure or texts in any other way than specified, or the non-experts will lose interest.
\\
\texttt{PI2} & Think of yourself as a translator: You're translating complex, technical language (found inside [square brackets]) into everyday language.
The Job: You will be given 10 documents all at the same time.
Your Task for Each Document:
    Find all [bracketed phrases].
    "Translate" them into simple terms for a general audience.
    Important: Do not "translate" or change anything outside the brackets. Keep the original sentence flow. Result Expected: A list of 10 translated documents, ensuring all original bracketed items have been simplified. Do not output the original input texts.
\\
    \end{tabular}
    \caption{Prompts used for the batches of ten texts.}
    \label{tab:prompts}
\end{table}

\begin{table}[t]
    \centering
    \begin{tabular}{l|r|r|p{7.5cm}}
        \textbf{Run ID} & \textbf{\# f10} &\textbf{\# f1} & \textbf{Description} \\ \hline
       \texttt{baseline} & 0 & 0 & Nothing changed, complex sentences = simplified sentences \\ \hline

        \texttt{c--gpt-4.1-nano} & 8 & 1 &  Make original texts more complex but not longer (prompt \texttt{C}) + usage of OpenAI \\
        \texttt{c--gemini-2.5-flash-preview} & 18 & 7 & Make original texts more complex but not longer (prompt \texttt{C}) + usage of Gemini 2.5 \\
        \texttt{c--gemini-2.0-flash} & 10 & 1 & Make original texts more complex but not longer (prompt \texttt{C}) + usage of Gemini 2.0\\ \hline

        \texttt{r--gemini-2.5-flash-preview} & 1 & 0 & Rephrase original texts with same complexity and length (prompt \texttt{R}) + usage of Gemini 2.5 \\
        \texttt{r--gemini-2.0-flash} & 47 & 1 & Rephrase original texts with same complexity and length (prompt \texttt{R}) + usage of Gemini 2.0\\ \hline

        \texttt{p1--gpt-4.1-nano} & 67 & 22 & Complex term identification + new prompt (\texttt{P1}) + usage of OpenAI\\
        \texttt{p1--gemini-2.0-flash} & 42 & 22 & Complex term identification + new prompt (\texttt{P1}) + usage of Gemini 2.0\\ 
        \texttt{p1-ac--gemini-2.0-flash} & 44 & 22 & \texttt{c--gpt-4.1-nano} complexification +  complex term identification + new prompt (\texttt{P1}) + usage of Gemini 2.0\\ 
\hline

        \texttt{p2--gpt-4.1-nano} & 75 & 21 & Complex term identification + old prompt (\texttt{P2}) + usage of OpenAI\\
        \texttt{p2--gemini-2.5-flash-preview} & 275 & 1550 & Complex term identification + old prompt (\texttt{P2}) + usage of Gemini 2.5 \\
        \texttt{p2--gemini-2.0-flash} & 87 & 132 & Complex term identification + old prompt (\texttt{P2}) + usage of Gemini 2.0\\
       \texttt{p2-ac--gemini-2.0-flash} & 91 & 132 & c--gpt-4.1-nano complexification +  complex term identification + old prompt (\texttt{P2}) + usage of Gemini 2.0\\ \hline

\texttt{pni1--gpt-4.1-nano}	& 32 & <320 & Complex term identification + new prompt (\texttt{PNI1}) + usage of OpenAI \\
\texttt{pn1--gemini-2.0-flash}	&  174 & 1268 & Complex term identification + new prompt (\texttt{PN1}) + usage of Gemini 2.0	\\
\texttt{pi2--gemini-2.0-flash}	& 199 & 174 & Complex term identification + new prompt (\texttt{PI2}) + usage of Gemini 2.0	\\

    \end{tabular}

    \caption{Information on run IDs, the number of failed modifications for the batch processing (\# f10), single processing (\# f1) and the description of the run. Our official Run IDs have a \texttt{THM\_task11\_}-prefix. Due to an error \# f1 of \texttt{pni1--gpt-4.1-nano} has not been tracked.}
    \label{tab:runs}
\end{table}

We investigate three steps which can all be optionally employed in text simplification. At first a \textit{rephrasing}, e.g., via an LLM, can modify the original text. Then for an original or modified text the \textit{complex scientific terms} can be identified and marked specifically. As a last step, the actual \textit{simplification} is run on original or modified texts, which either have complex terms marked or not, e.g., by again using an LLM.

\subsection{Step 1: Rephrasing}

For runs incorporating this step, we employ a prompt that reformulates the original text. This can be to make an original text \textit{more complex} (see prompt \texttt{C} in \autoref{tab:prompts}) or to simply \textit{rephrase} it while striving to maintain complexity and length (see prompt \texttt{P} in \autoref{tab:prompts}). In general we experiment in three directions: modifying the original text, introducing more complexity and simplifying it immediately. The attempt to increase a text's complexity is assumed to produce worse evaluation results if the texts are not at their maximum complexity level yet. If the texts' complexity is already the highest it can be, an experiment trying to complexify the text will lead to results that are comparable to rephrasing.

\subsection{Step 2: Complex Scientific Term Identification}

For runs incorporating this step, the general idea is to identify the complex scientific terms in the original or rephrased texts and specifically mark the parts that should be replaced in the simplification step.

We run the complex term identification as implemented by team IRGC'23~\cite{DBLP:conf/clef/0002HKNS23}. They mark a subset of complex terms identified by an established but highly sensitive keyphrase extraction method~\cite{kulkarni-etal-2022-learning}. The intuition behind this is that not all keyphrases are complex and not all complex keyphrases are difficult due to domain jargon.
Idf scores for terms are calculated based on two corpora which we use through PyTerrier~\cite{10.1145/3409256.3409829}: \texttt{lotte/lifestyle}, a corpus focused on texts from lifestyle forums and \texttt{lotte/science}, a corpus focused on texts from science-oriented forums~\cite{santhanam2022colbertv2effectiveefficientretrieval}. Terms with a higher idf in science texts than in lifestyle texts are considered to be domain jargon which should be explained to non-experts. 
Then, only extracted keyphrases where their idf surpasses a certain complexity threshold are marked as complex terms which should be explained to non-experts. In our experiments we work with a complexity threshold of 0.1\footnote{IRGC'23 used a complexity threshold of 0.01 which is more sensitive whereas our usage of 0.1 is more strict in identifying complex terms in the science domain, supposedly reducing the number of false positive complex terms.}.

From the 9160 original texts in the dataset, 3473 texts did not receive any markings indicating complex phrases.

\subsection{Step 3: Simplification}

All our prompts ask LLMs to take on a specific persona to simplify the texts as this prompting technique is highly effective in numerous tasks~\cite{white2023promptpatterncatalogenhance}.
For simplifying the texts, we run our prompts (see prompts \texttt{P1}, \texttt{P2}, \texttt{PNI1}, \texttt{PN1} and \texttt{PI2} in \autoref{tab:prompts}).
While \texttt{P2} is the prompt introduced by IRGC'23~\cite{DBLP:conf/clef/0002HKNS23} defining the LLM's expected persona not specifically, \texttt{P1} clearly states to behave as a text simplification system where no domain-knowledge should be assumed in the simplification. 
\texttt{PNI1} treats the LLM more informally and similar to a human but does imply the expected persona by stating it being the \textit{best and most reliable system for text simplification and language in general}. 
\texttt{PN1} defines the LLM's persona as someone attending a conference.
\texttt{PI2} instructs the LLM to act as a translator translating complex technical language to everyday language.

In general we run our prompts for randomly batched texts of size 10, if prompts do not return 10 modified texts, we run these texts again one on one in the end with slightly modified versions of the prompts to work on single texts instead of 10. If these single runs also do not produce single texts as output, we consider the original (complex) text as the simplified text as well.

\section{Runs and Results}

\subsection{Submitted Runs}
\begin{table}[t]
    \centering
    \begin{tabular}{l|p{13.25cm}}
        \textbf{Processing} & \textbf{Prompt text} \\ \hline
         Batch & You are given 10 texts, TREAT THEM SEPARATELY.
Complex technical or scientific terms are indicated by square brackets (e.g. [convolutional neural network]).
For each text, replace or explain the terms in square brackets to make it understandable to non-experts and to simplify for non-expert readers.
Make sure that you return exactly one enumerated list of 10 simplified texts and that all terms in square brackets are simplified or explained!\\

        Single & You are given 1 text.
Complex technical or scientific terms are indicated by square brackets (e.g. [convolutional neural network]).
For each text, replace or explain the terms in square brackets to make it understandable to non-experts and to simplify for non-expert readers.
Make sure that you return exactly one enumerated list of 1 simplified text and that all terms in square brackets are simplified or explained!
\\
         
    \end{tabular}
    \caption{Comparison of the batch processing prompt of \texttt{P2} with its single text processing variant.}
    \label{tab:prompt_single}
\end{table}

Our submitted runs have the prefix of our team name \texttt{THM} combined with the task number, prompt ID as well as used LLM, resulting in run names such as \texttt{THM\_task11\_p1-gpt-4.1-nano} with \texttt{p1} being a prompt ID and \texttt{gpt-4.1-nano} describing the used model in a run. For readability we refrain from writing out the first part of the run IDs (\texttt{THM\_task11\_}) in this paper.
For our simplification prompts we defined the batch processing versions in \autoref{tab:prompts}, the single text versions are almost identical, \autoref{tab:prompt_single} holds an example for \texttt{P2}.

\autoref{tab:runs} gives an overview of our submitted runs as well as the number of cases in which we did not receive the expected number of simplifications when using an LLM.
We experiment with usage of small Gemini models (\texttt{gemini-2.0-flash} and \texttt{gemini-2.5-preview-flash}) and a small OpenAI model (\texttt{gpt-4.1-nano}).
Our implementation can be found on GitHub\footnote{\url{https://github.com/kreutzch/THM-SimpleText-25}}.

\subsection{Results}
\begin{table}[]
    \centering
    \begin{tabular}{l|r|r}
        \textbf{Run ID} & \textbf{SARI simple original} & \textbf{SARI simple auto}\\ \hline
\texttt{baseline}	&	7,844	&	12.033	\\
\hline
\texttt{c--gpt-4.1-nano}	&	32.445	&	33.944	\\
\texttt{c--gemini-2.5-flash-preview}	&	30.175	&	32.586	\\
\texttt{c--gemini-2.0-flash}	&	28.917	&	31.199	\\

\hline
\texttt{r--gemini-2.5-flash-preview}	&	29.256	&	32.598	\\ 
\texttt{r--gemini-2.0-flash}	&	27.526	&	31.160	\\
\hline

\texttt{p1--gpt-4.1-nano}	&	38.238	&	40.416	\\
\texttt{p1--gemini-2.0-flash}	&	25.858	&	30.001	\\
\texttt{p1-ac--gemini-2.0-flash} & 25.234 & 29.080 \\

\hline
\texttt{p2--gpt-4.1-nano}	&	\textbf{39.572}	&	\textbf{41.315}	\\
\texttt{p2--gemini-2.5-flash-preview}	&	25.062	&	27.811	\\
\texttt{p2--gemini-2.0-flash}	&	29.263	&	31.174	\\
\texttt{p2-ac--gemini-2.0-flash}	&	27.561	&	30.024	\\

\hline
\texttt{pni1--gpt-4.1-nano}	&	35.262	&	37.262	\\
\texttt{pn1--gemini-2.0-flash}	&	32.267	&	34.474	\\
\texttt{pi2--gemini-2.0-flash}	&	20.521	&	24.045	\\

    \end{tabular}
    \caption{SARI~\cite{sari} scores of submitted runs produced by \href{https://www.codabench.org/competitions/8400/}{Codabench}.}
    \label{tab:results}
\end{table}

\begin{table}[t]
    \centering
    \footnotesize
    \begin{tabular}{l|l|p{3.7cm}|p{3.7cm}|p{3.7cm}}
        \textbf{Para ID} & \textbf{Sent ID} & \textbf{Original} & \textbf{Simplified with \texttt{P1}} & \textbf{Simplified with \texttt{P2}}\\ \hline
0 & 5& Quality of care outcomes (proportions of patients receiving evidence-based care) were included in all included studies. & 
\textcolor{green}{All the studies included looked at the} quality of care\textcolor{green}{, meaning how many} patients receiv\textcolor{green}{ed care based on the best available evidence.} & \textcolor{green}{The} quality of care outcomes (\textcolor{green}{how many} patients receiv\textcolor{green}{ed care based on the best available evidence}) were included in all \textcolor{green}{the} studies.\\

5 & 10 & For every nine people treated with haloperidol instead of olanzapine, one fewer person would experience clinically important improvement in quality of life. & 
For every nine people treated with haloperidol \textcolor{green}{(a medication used for mental health conditions)} instead of olanzapine \textcolor{green}{(another mental health drug)}, one fewer person would experience \textcolor{green}{a meaningful} improvement in \textcolor{green}{their} quality of life. & 
For every nine people treated with haloperidol \textcolor{green}{(a medication used for mental health conditions)} instead of olanzapine \textcolor{green}{(another mental health drug)}, one fewer person would experience \textcolor{green}{a meaningful} improvement in quality of life.\\

6& 18& Telephone interventions provide a convenient way of supporting self-management of cancer-related symptoms for adults with cancer. & 
Telephone\textcolor{green}{-based programs offer} a convenient way \textcolor{green}{to} support adults with cancer \textcolor{green}{in managing symptoms related to their illness on their own.}& 
Telephone\textcolor{green}{-based programs offer} a convenient way \textcolor{green}{to} support \textcolor{green}{people in managing symptoms related to cancer on their own.}\\
    \end{tabular}
    \caption{Examples of original texts as well as texts produced with prompts \texttt{P1} and \texttt{P2} with \texttt{gpt-4.1-nano}. Parts marked in green signal the modifications compared to the original text.}
    \label{tab:examples_p1_p2}
\end{table}

We found considerable differences between employing \texttt{gpt-4.1-nano} and the two versions of Gemini with the OpenAI model producing better results in all cases. Usage of the more powerful model \texttt{gemini-2.5-flash-preview} did not consistently produce better scores compared to the older model \texttt{gemini-2.0-flash} (see prompt \texttt{P2}).
All considered different configuration of runs using prompt \texttt{P2} produced better results than runs with the same configurations using prompt \texttt{P1}. \texttt{P1} is the more descriptive version of the more concise \texttt{P2}. In \texttt{P1} the context of being a text simplification system with definition of the knowledge is explicitly defined. Additionally, \texttt{P1} puts extra emphasis on the requirement of returning only 10 texts while keeping the structure of the original text. We assume the longer prompt with more instructions dilutes the LLM's attention from the actual task of simplifying texts.
\autoref{tab:examples_p1_p2} gives some randomly selected examples of original texts as well as their simplifications using \texttt{P1} and \texttt{P2} combined with \texttt{gpt-4.1-nano}. It appears that both prompts \textit{produced almost the same simplification} in some cases (as seen in the second example), \texttt{P2} \textit{staying closer to the original text} (as seen in the first example) and \texttt{P2} producing slightly \textit{shorter or more concise} texts (as seen in the third example).
Both prompts do not seem to fully convey the instruction of only modifying the identified complex terms in original texts.

\begin{table}[t]
    \centering

    \footnotesize
    \begin{tabular}{l|l|p{3.7cm}|p{3.7cm}|p{3.7cm}}
        \textbf{Para ID} & \textbf{Sent ID} & \textbf{Original} & \textbf{Rephrased with \texttt{R}} & \textbf{Complexified with \texttt{C}}\\ \hline
0 & 5 & Quality of care outcomes (proportions of patients receiving evidence-based care) were included in all included studies. & All studies incorporated quality of care outcomes, which measured the proportion of patients receiving evidence-based care. & Parameters indicative of care quality (specifically, the prevalence of patients undergoing empirically validated therapeutic interventions) were incorporated into all pertinent scholarly endeavors.\\

5 & 10 & For every nine people treated with haloperidol instead of olanzapine, one fewer person would experience clinically important improvement in quality of life. & If nine individuals were treated with haloperidol instead of olanzapine, one fewer person would achieve a clinically significant improvement in their quality of life. & For every nine individuals receiving haloperidol in lieu of olanzapine, a reduction of one individual would be observed in those experiencing a clinically significant amelioration in their overall quality of life.\\

6 & 18 & Telephone interventions provide a convenient way of supporting self-management of cancer-related symptoms for adults with cancer. & Telephone interventions offer a practical method for assisting adults with cancer in self-managing their cancer-related symptoms. & 6. Telephonic interventions offer an accessible modality for facilitating autonomous symptom navigation of oncological sequelae for adult oncological patients.\\
    \end{tabular}
    \caption{Examples of original texts as well as texts produced with prompts \texttt{R} and \texttt{C} with \texttt{gemini-2.5-flash-preview}.}
    \label{tab:examples}
\end{table}

\autoref{tab:results} indicates that the runs instructing the LLMs to complexify the original sentences did not perform worse than a rephrasing of original sentences without reducing complexity or length.
Table~\autoref{tab:examples} gives again gives the same three examples as seen in \autoref{tab:examples_p1_p2} of original texts with their reprasing as well as their complexification. It appears as if complexified sentences remained closer to the original versions than the rephrased ones which could lead at increased SARI scores as they are also closer to the gold simplification.
This hints at problems with the employed evaluation measures.

\begin{table}[t]
    \centering
    \footnotesize
    \begin{tabular}{l|p{13.75cm}}
        \textbf{Prompt} & \textbf{Simplification} \\ \hline

\texttt{PNI1} & For every nine people treated with haloperidol \textcolor{green}{(a medication for mental health conditions)} instead of olanzapine \textcolor{green}{(another mental health medication), }one fewer person would \textcolor{green}{see a meaningful} improvement in \textcolor{green}{their} quality of life.\\

\texttt{PN1} & \textcolor{green}{Okay, everyone, let's look at this statement. It's comparing two different medications, haloperidol and olanzapine. Think of haloperidol as an older type of medication used to treat certain mental health conditions. The statement is saying that if you treat} nine people \textcolor{green}{with this older medication,} haloperidol\textcolor{green}{,} instead of olanzapine\textcolor{green}{, which is a newer medication, you'll find that *one*} fewer person \textcolor{green}{will see a meaningful improvement in their overall well-being or "}quality of life.\textcolor{green}{" So, olanzapine seems to be giving people a better chance at improving their quality of life compared to haloperidol, according to this statement.}\\

\texttt{PI2} & For every nine people treated with \textcolor{green}{a strong, older antipsychotic medication} instead of olanzapine, one fewer person would experience clinically important improvement in quality of life.\\
         
    \end{tabular}
    \caption{Comparison of simplifications for prompts \texttt{PNI1}, \texttt{PN1} and \texttt{PI2} for \textbf{Para ID} = 5 and \textbf{Sent ID} = 10. Parts marked in green signal the modifications compared to the original text.}
    \label{tab:bad_prompts}
\end{table}

\autoref{tab:bad_prompts} gives an example for a simplification produced by prompts \texttt{PNI1}, \texttt{PN1} and \texttt{PI2}. Here we see the closeness of the text produced by \texttt{PNI1} to the original text while explaining the complex terms, the complete rephrasing of the text produced by \texttt{PN1} that also simplified the original text significantly and the text produced by \texttt{PI2} that did keep a good chunk of the original text and did not simplify it much. Failure to simplify the complex parts of the original texts does unsurprisingly lead to lower SARI scores.

From our experiments, \autoref{tab:runs} and \autoref{tab:results} we conclude the description of the return type specified to a LLM as well as the post-processing of the result being highly influential of the overall quality of the run. Runs utilising the Gemini variants produced higher error rates than the GPT runs with the exception of \texttt{P1}. In our experiments we felt that it was more difficult to write prompts suitable for Gemini for this exact reason.

\subsection{Costs}

\begin{table}[t]
    \centering
    \begin{tabular}{l|r|r|r|r}
        \textbf{Model} & \textbf{\$ Input} & \textbf{\$ Output} & \textbf{\# Input Tokens} & \textbf{\# Requests}\\\hline
        \texttt{gpt-4.1-nano} & \$1.072 & \$3.295 & 10.833M & 30,975\\
        \texttt{gpt-3.5-turbo-0125}  & \$0.833 & \$0.602 & 1.666M & 2,746\\

        \texttt{gemini-2.5-flash-preview} & \$0.45 & \$1.44 &  3.655M & 9.09k\\
        \texttt{gemini-2.0-flash} & \$0.75 & \$2.85& 9.93M & 21.39k\\

    \end{tabular}
    \caption{Overview of cost in \$ for used LLMs during all experiments and runs with input number of tokens and requests.}
    \label{tab:llm_cost}
\end{table}

\autoref{tab:llm_cost} provides an overview of cost, input tokens and requests per model.

At the time of producing the runs, the used OpenAI model \texttt{gpt-4.1-nano} cost \$0.10 per 1M tokens. 
We also experimented with usage of \texttt{gpt-3.5-turbo-0125} with a cost of \$0.50 per 1M tokens. 
During experimentation as well as running our actual prompts we issued 33,721 requests with 12.499M input tokens. In total we spent \$6.39 using OpenAI models.

Both our used Gemini models \texttt{gemini-2.5-flash-preview} and \texttt{gemini-2.0-flash} cost \$0.10 per 1M input tokens. For both models, the output cost was \$0.40 per 1M tokens.
In total we issued 30.48k requests with 13.585M input tokens. The total costs for all models used was \$5.49. 

The full cost of using LLMs for our participation in the shared task was thus \$11.88. We issued 64.2k requests with 26.084M input tokens.

\section{Conclusion}

Our experiments show the potential value of combining a cheap complex term identification step with low-cost LLM options for simplifying complex texts for non-expert readers. It appears prompts producing texts that remain more similar to the original texts perform better in the evaluation compared to prompts producing texts with more changed up wording. The actual difficulty of produced texts does not seem to be the sole determining factor of the goodness of the strategy. 

Future work should focus on experimenting with locally-run LLMs and other options to identify complex keyphrases in texts. Another line of work should research better automatic evaluation measures for the text simplification task.

\section*{Acknowledgments}
We thank the SimpleText organisers for continuously organising the shared task and bringing together the community to advance the field of automatic text simplification.

\section*{Declaration on Generative AI}

The authors have not employed any Generative AI tools for writing the manuscript.
The authors used OpenAI and Gemini models in their implementation.

\bibliography{sample-ceur}

\begin{thebibliography}{19}
\expandafter\ifx\csname natexlab\endcsname\relax\def\natexlab#1{#1}\fi
\providecommand{\url}[1]{\texttt{#1}}
\providecommand{\href}[2]{#2}
\providecommand{\path}[1]{#1}
\providecommand{\DOIprefix}{doi:}
\providecommand{\ArXivprefix}{arXiv:}
\providecommand{\URLprefix}{URL: }
\providecommand{\Pubmedprefix}{pmid:}
\providecommand{\doi}[1]{\href{http://dx.doi.org/#1}{\path{#1}}}
\providecommand{\Pubmed}[1]{\href{pmid:#1}{\path{#1}}}
\providecommand{\bibinfo}[2]{#2}
\ifx\xfnm\relax \def\xfnm[#1]{\unskip,\space#1}\fi
\bibitem[{Ermakova et~al.(2021)Ermakova, Bellot, Braslavski, Kamps, Mothe, Nurbakova, Ovchinnikova, and SanJuan}]{DBLP:conf/ecir/ErmakovaBBKMNOS21}
\bibinfo{author}{L.~Ermakova}, \bibinfo{author}{P.~Bellot}, \bibinfo{author}{P.~Braslavski}, \bibinfo{author}{J.~Kamps}, \bibinfo{author}{J.~Mothe}, \bibinfo{author}{D.~Nurbakova}, \bibinfo{author}{I.~Ovchinnikova}, \bibinfo{author}{E.~SanJuan},
\newblock \bibinfo{title}{Text simplification for scientific information access - {CLEF} 2021 simpletext workshop},
\newblock in: \bibinfo{booktitle}{Advances in Information Retrieval - 43rd European Conference on {IR} Research, {ECIR} 2021, Virtual Event, March 28 - April 1, 2021, Proceedings, Part {II}}, volume \bibinfo{volume}{12657}, \bibinfo{year}{2021}, pp. \bibinfo{pages}{583--592}. \DOIprefix\doi{10.1007/978-3-030-72240-1\_68}.
\bibitem[{Ermakova et~al.(2022)Ermakova, Bellot, Kamps, Nurbakova, Ovchinnikova, SanJuan, Mathurin, Ara{\'{u}}jo, Hannachi, Huet, and Poinsu}]{DBLP:conf/ecir/ErmakovaBKNOSMA22}
\bibinfo{author}{L.~Ermakova}, \bibinfo{author}{P.~Bellot}, \bibinfo{author}{J.~Kamps}, \bibinfo{author}{D.~Nurbakova}, \bibinfo{author}{I.~Ovchinnikova}, \bibinfo{author}{E.~SanJuan}, \bibinfo{author}{{\'{E}}.~Mathurin}, \bibinfo{author}{S.~Ara{\'{u}}jo}, \bibinfo{author}{R.~Hannachi}, \bibinfo{author}{S.~Huet}, \bibinfo{author}{N.~Poinsu},
\newblock \bibinfo{title}{Automatic simplification of scientific texts: Simpletext lab at {CLEF-2022}},
\newblock in: \bibinfo{booktitle}{Advances in Information Retrieval - 44th European Conference on {IR} Research, {ECIR} 2022, Stavanger, Norway, April 10-14, 2022, Proceedings, Part {II}}, volume \bibinfo{volume}{13186}, \bibinfo{year}{2022}, pp. \bibinfo{pages}{364--373}. \DOIprefix\doi{10.1007/978-3-030-99739-7\_46}.
\bibitem[{Ermakova et~al.(2023)Ermakova, SanJuan, Huet, Augereau, Azarbonyad, and Kamps}]{DBLP:conf/ecir/ErmakovaSHAAK23}
\bibinfo{author}{L.~Ermakova}, \bibinfo{author}{E.~SanJuan}, \bibinfo{author}{S.~Huet}, \bibinfo{author}{O.~Augereau}, \bibinfo{author}{H.~Azarbonyad}, \bibinfo{author}{J.~Kamps},
\newblock \bibinfo{title}{{CLEF} 2023 simpletext track - what happens if general users search scientific texts?},
\newblock in: \bibinfo{booktitle}{Advances in Information Retrieval - 45th European Conference on Information Retrieval, {ECIR} 2023, Dublin, Ireland, April 2-6, 2023, Proceedings, Part {III}}, volume \bibinfo{volume}{13982}, \bibinfo{year}{2023}, pp. \bibinfo{pages}{536--545}. \DOIprefix\doi{10.1007/978-3-031-28241-6\_62}.
\bibitem[{Ermakova et~al.(2024)Ermakova, SanJuan, Huet, Azarbonyad, Nunzio, Vezzani, D'Souza, Kabongo, Giglou, Zhang, Auer, and Kamps}]{DBLP:conf/ecir/ErmakovaSHANVDKGZAK24}
\bibinfo{author}{L.~Ermakova}, \bibinfo{author}{E.~SanJuan}, \bibinfo{author}{S.~Huet}, \bibinfo{author}{H.~Azarbonyad}, \bibinfo{author}{G.~M.~D. Nunzio}, \bibinfo{author}{F.~Vezzani}, \bibinfo{author}{J.~D'Souza}, \bibinfo{author}{S.~Kabongo}, \bibinfo{author}{H.~B. Giglou}, \bibinfo{author}{Y.~Zhang}, \bibinfo{author}{S.~Auer}, \bibinfo{author}{J.~Kamps},
\newblock \bibinfo{title}{{CLEF} 2024 simpletext track - improving access to scientific texts for everyone},
\newblock in: \bibinfo{booktitle}{Advances in Information Retrieval - 46th European Conference on Information Retrieval, {ECIR} 2024, Glasgow, UK, March 24-28, 2024, Proceedings, Part {VI}}, volume \bibinfo{volume}{14613}, \bibinfo{year}{2024}, pp. \bibinfo{pages}{28--35}. \DOIprefix\doi{10.1007/978-3-031-56072-9\_4}.
\bibitem[{Ermakova et~al.(2025{\natexlab{a}})Ermakova, Azarbonyad, Bakker, Vendeville, and Kamps}]{DBLP:conf/ecir/ErmakovaABVK25}
\bibinfo{author}{L.~Ermakova}, \bibinfo{author}{H.~Azarbonyad}, \bibinfo{author}{J.~Bakker}, \bibinfo{author}{B.~Vendeville}, \bibinfo{author}{J.~Kamps},
\newblock \bibinfo{title}{{CLEF} 2025 simpletext track - simplify scientific text (and nothing more)},
\newblock in: \bibinfo{booktitle}{Advances in Information Retrieval - 47th European Conference on Information Retrieval, {ECIR} 2025, Lucca, Italy, April 6-10, 2025, Proceedings, Part {V}}, volume \bibinfo{volume}{15576}, \bibinfo{year}{2025}{\natexlab{a}}, pp. \bibinfo{pages}{425--433}. \DOIprefix\doi{10.1007/978-3-031-88720-8\_63}.
\bibitem[{Ermakova et~al.(2025{\natexlab{b}})Ermakova, Azarbonyad, Bakker, Vendeville, and Kamps}]{simpletext-lncs}
\bibinfo{author}{L.~Ermakova}, \bibinfo{author}{H.~Azarbonyad}, \bibinfo{author}{J.~Bakker}, \bibinfo{author}{B.~Vendeville}, \bibinfo{author}{J.~Kamps},
\newblock \bibinfo{title}{Overview of the {CLEF 2025 SimpleText} track: Simplify scientific texts (and nothing more)},
\newblock in: \bibinfo{editor}{J.~Carrillo~de Albornoz}, \bibinfo{editor}{J.~Gonzalo}, \bibinfo{editor}{L.~Plaza}, \bibinfo{editor}{A.~García Seco~de Herrera}, \bibinfo{editor}{J.~Mothe}, \bibinfo{editor}{F.~Piroi}, \bibinfo{editor}{P.~Rosso}, \bibinfo{editor}{D.~Spina}, \bibinfo{editor}{G.~Faggioli}, \bibinfo{editor}{N.~Ferro} (Eds.), \bibinfo{booktitle}{Experimental IR Meets Multilinguality, Multimodality, and Interaction. Proceedings of the Sixteenth International Conference of the CLEF Association (CLEF 2025)}, Lecture Notes in Computer Science, \bibinfo{publisher}{Springer}, \bibinfo{year}{2025}{\natexlab{b}}.
\bibitem[{Kreutz et~al.(2024)Kreutz, Haak, Engelmann, and Schaer}]{DBLP:conf/acl/KreutzH0S24}
\bibinfo{author}{C.~Kreutz}, \bibinfo{author}{F.~Haak}, \bibinfo{author}{B.~Engelmann}, \bibinfo{author}{P.~Schaer},
\newblock \bibinfo{title}{{BATS:} benchmarking text simplicity},
\newblock in: \bibinfo{booktitle}{Findings of the Association for Computational Linguistics, {ACL} 2024, Bangkok, Thailand and virtual meeting, August 11-16, 2024}, \bibinfo{year}{2024}, pp. \bibinfo{pages}{11968--11989}. \DOIprefix\doi{10.18653/V1/2024.FINDINGS-ACL.712}.
\bibitem[{{\v{S}}tajner and Hulpu{\c{s}}(2018)}]{stajner-hulpus-2018-automatic}
\bibinfo{author}{S.~{\v{S}}tajner}, \bibinfo{author}{I.~Hulpu{\c{s}}},
\newblock \bibinfo{title}{Automatic assessment of conceptual text complexity using knowledge graphs},
\newblock in: \bibinfo{booktitle}{Proceedings of the 27th International Conference on Computational Linguistics}, \bibinfo{year}{2018}, pp. \bibinfo{pages}{318--330}. \URLprefix \url{https://aclanthology.org/C18-1027/}.
\bibitem[{{\v{S}}tajner et~al.(2021){\v{S}}tajner, Nisioi, and Ibanez}]{10.1145/3439231.3439263}
\bibinfo{author}{S.~{\v{S}}tajner}, \bibinfo{author}{S.~Nisioi}, \bibinfo{author}{D.~Ibanez},
\newblock \bibinfo{title}{Is simple english wikipedia as simple and easy-to-understand as we expect it to be?},
\newblock in: \bibinfo{booktitle}{Proceedings of the 9th International Conference on Software Development and Technologies for Enhancing Accessibility and Fighting Info-Exclusion}, DSAI '20, \bibinfo{year}{2021}, p. \bibinfo{pages}{66–70}. \DOIprefix\doi{10.1145/3439231.3439263}.
\bibitem[{Padovani et~al.(2024)Padovani, Marchesi, Pasqua, Galletti, and Nardi}]{padovani-etal-2024-automatic}
\bibinfo{author}{F.~Padovani}, \bibinfo{author}{C.~Marchesi}, \bibinfo{author}{E.~Pasqua}, \bibinfo{author}{M.~Galletti}, \bibinfo{author}{D.~Nardi},
\newblock \bibinfo{title}{Automatic text simplification: A comparative study in {I}talian for children with language disorders},
\newblock in: \bibinfo{booktitle}{Proceedings of the 13th Workshop on Natural Language Processing for Computer Assisted Language Learning}, \bibinfo{year}{2024}, pp. \bibinfo{pages}{176--186}. \URLprefix \url{https://aclanthology.org/2024.nlp4call-1.13/}.
\bibitem[{{\v{S}}tajner(2021)}]{stajner-2021-automatic}
\bibinfo{author}{S.~{\v{S}}tajner},
\newblock \bibinfo{title}{Automatic text simplification for social good: Progress and challenges},
\newblock in: \bibinfo{booktitle}{Findings of the Association for Computational Linguistics: ACL-IJCNLP 2021}, \bibinfo{year}{2021}, pp. \bibinfo{pages}{2637--2652}. \DOIprefix\doi{10.18653/v1/2021.findings-acl.233}.
\bibitem[{Gooding(2022)}]{gooding-2022-ethical}
\bibinfo{author}{S.~Gooding},
\newblock \bibinfo{title}{On the ethical considerations of text simplification},
\newblock in: \bibinfo{editor}{S.~Ebling}, \bibinfo{editor}{E.~Prud{'}hommeaux}, \bibinfo{editor}{P.~Vaidyanathan} (Eds.), \bibinfo{booktitle}{Ninth Workshop on Speech and Language Processing for Assistive Technologies (SLPAT-2022)}, \bibinfo{year}{2022}, pp. \bibinfo{pages}{50--57}. \DOIprefix\doi{10.18653/v1/2022.slpat-1.7}.
\bibitem[{Bakker et~al.(2025)Bakker, Vendeville, Ermakova, and Kamps}]{simpletext-1}
\bibinfo{author}{J.~Bakker}, \bibinfo{author}{B.~Vendeville}, \bibinfo{author}{L.~Ermakova}, \bibinfo{author}{J.~Kamps},
\newblock \bibinfo{title}{{Overview of the CLEF 2025 SimpleText Task 1: Simplify Scientific Text}},
\newblock in: \bibinfo{editor}{G.~Faggioli}, \bibinfo{editor}{N.~Ferro}, \bibinfo{editor}{P.~Rosso}, \bibinfo{editor}{D.~Spina} (Eds.), \bibinfo{booktitle}{Working Notes of {CLEF} 2025: Conference and Labs of the Evaluation Forum}, {CEUR} Workshop Proceedings, \bibinfo{publisher}{CEUR-WS.org}, \bibinfo{year}{2025}.
\bibitem[{Engelmann et~al.(2023)Engelmann, Haak, Kreutz, Nikzad{-}Khasmakhi, and Schaer}]{DBLP:conf/clef/0002HKNS23}
\bibinfo{author}{B.~Engelmann}, \bibinfo{author}{F.~Haak}, \bibinfo{author}{C.~K. Kreutz}, \bibinfo{author}{N.~Nikzad{-}Khasmakhi}, \bibinfo{author}{P.~Schaer},
\newblock \bibinfo{title}{Text simplification of scientific texts for non-expert readers},
\newblock in: \bibinfo{booktitle}{Working Notes of the Conference and Labs of the Evaluation Forum {(CLEF} 2023), Thessaloniki, Greece, September 18th to 21st, 2023}, volume \bibinfo{volume}{3497}, \bibinfo{year}{2023}, pp. \bibinfo{pages}{2987--2998}. \URLprefix \url{https://ceur-ws.org/Vol-3497/paper-250.pdf}.
\bibitem[{Kulkarni et~al.(2022)Kulkarni, Mahata, Arora, and Bhowmik}]{kulkarni-etal-2022-learning}
\bibinfo{author}{M.~Kulkarni}, \bibinfo{author}{D.~Mahata}, \bibinfo{author}{R.~Arora}, \bibinfo{author}{R.~Bhowmik},
\newblock \bibinfo{title}{Learning rich representation of keyphrases from text},
\newblock in: \bibinfo{booktitle}{Findings of the Association for Computational Linguistics: NAACL 2022}, \bibinfo{year}{2022}, pp. \bibinfo{pages}{891--906}. \DOIprefix\doi{10.18653/v1/2022.findings-naacl.67}.
\bibitem[{Macdonald and Tonellotto(2020)}]{10.1145/3409256.3409829}
\bibinfo{author}{C.~Macdonald}, \bibinfo{author}{N.~Tonellotto},
\newblock \bibinfo{title}{Declarative experimentation in information retrieval using pyterrier},
\newblock in: \bibinfo{booktitle}{Proceedings of the 2020 ACM SIGIR on International Conference on Theory of Information Retrieval}, ICTIR '20, \bibinfo{year}{2020}, p. \bibinfo{pages}{161–168}. \DOIprefix\doi{10.1145/3409256.3409829}.
\bibitem[{Santhanam et~al.(2022)Santhanam, Khattab, Saad-Falcon, Potts, and Zaharia}]{santhanam2022colbertv2effectiveefficientretrieval}
\bibinfo{author}{K.~Santhanam}, \bibinfo{author}{O.~Khattab}, \bibinfo{author}{J.~Saad-Falcon}, \bibinfo{author}{C.~Potts}, \bibinfo{author}{M.~Zaharia}, \bibinfo{title}{{C}ol{BERT}v2: Effective and efficient retrieval via lightweight late interaction}, \bibinfo{year}{2022}. \DOIprefix\doi{10.18653/v1/2022.naacl-main.272}.
\bibitem[{White et~al.(2023)White, Fu, Hays, Sandborn, Olea, Gilbert, Elnashar, Spencer-Smith, and Schmidt}]{white2023promptpatterncatalogenhance}
\bibinfo{author}{J.~White}, \bibinfo{author}{Q.~Fu}, \bibinfo{author}{S.~Hays}, \bibinfo{author}{M.~Sandborn}, \bibinfo{author}{C.~Olea}, \bibinfo{author}{H.~Gilbert}, \bibinfo{author}{A.~Elnashar}, \bibinfo{author}{J.~Spencer-Smith}, \bibinfo{author}{D.~C. Schmidt},
\newblock \bibinfo{title}{A prompt pattern catalog to enhance prompt engineering with chatgpt},
\newblock in: \bibinfo{booktitle}{Proceedings of the 30th Conference on Pattern Languages of Programs}, \bibinfo{year}{2023}. \DOIprefix\doi{10.5555/3721041.3721046}.
\bibitem[{Xu et~al.(2016)Xu, Napoles, Pavlick, Chen, and Callison-Burch}]{sari}
\bibinfo{author}{W.~Xu}, \bibinfo{author}{C.~Napoles}, \bibinfo{author}{E.~Pavlick}, \bibinfo{author}{Q.~Chen}, \bibinfo{author}{C.~Callison-Burch},
\newblock \bibinfo{title}{Optimizing statistical machine translation for text simplification},
\newblock \bibinfo{journal}{Transactions of the Association for Computational Linguistics} \bibinfo{volume}{4} (\bibinfo{year}{2016}) \bibinfo{pages}{401--415}. \DOIprefix\doi{10.1162/tacl_a_00107}.

\end{thebibliography}


\end{document}